\documentclass[conference]{IEEEtran}
\IEEEoverridecommandlockouts
\usepackage{cite}
\usepackage{amsmath,amssymb,amsfonts}
\usepackage{algorithmic}
\usepackage{graphicx}
\usepackage{textcomp}
\usepackage{xcolor}
\def\BibTeX{{\rm B\kern-.05em{\sc i\kern-.025em b}\kern-.08em
    T\kern-.1667em\lower.7ex\hbox{E}\kern-.125emX}}
\begin{document}

\title{Improvement and Implementation of a Speech Emotion Recognition Model Based on Dual-Layer LSTM\\
\thanks{Identify applicable funding agency here. If none, delete this.}
}

\author{\IEEEauthorblockN{Xiaoran Yang}
\IEEEauthorblockA{\textit{Dept Computer Science} \\
\textit{Communication University of China}\\
Raleigh, United States\\
im.yxr.cs@gmail.com}
\and
\IEEEauthorblockN{Shuhan Yu}
\IEEEauthorblockA{\textit{Hainan International College} \\
\textit{Communication University of China}\\
Hainan, China \\
yushuhan1018@gmail.com}
\and
\IEEEauthorblockN{Wenxi Xu}
\IEEEauthorblockA{\textit{School of Economics} \\
\textit{Hefei University of Technology}\\
Hefei, China \\
xuwenxi040929@163.com}
}

\maketitle

\begin{abstract}
This paper builds upon an existing speech emotion recognition model by adding an additional LSTM layer to improve the accuracy and processing efficiency of emotion recognition from audio data. By capturing the long-term dependencies within audio sequences through a dual-layer LSTM network, the model can recognize and classify complex emotional patterns more accurately. Experiments conducted on the RAVDESS dataset validated this approach, showing that the modified dual-layer LSTM model improves accuracy by 2\% compared to the single-layer LSTM while significantly reducing recognition latency, thereby enhancing real-time performance. These results indicate that the dual-layer LSTM architecture is highly suitable for handling emotional features with long-term dependencies, providing a viable optimization for speech emotion recognition systems. This research provides a reference for practical applications in fields like intelligent customer service, sentiment analysis, and human-computer interaction.

\end{abstract}


\section{Introduction}
Speech Emotion Recognition (SER) is a core technology in artificial intelligence and human-computer interaction, aiming to recognize the emotional states of speakers by analyzing and processing audio signals. With the rising demand in applications such as intelligent customer service, emotional robots, and personalized recommendation systems, SER technology has gained widespread attention. The effectiveness of emotion recognition determines the naturalness of human-computer interaction and user experience, making it essential to improve SER model accuracy and real-time performance.

Traditional emotion recognition methods often rely on handcrafted features, such as Mel-Frequency Cepstral Coefficients (MFCC), pitch, and rhythm. However, these features cannot fully capture complex emotional information, especially when processing long-term dependencies in emotional features, where handcrafted features have limited expressive power. With the rapid advancement of deep learning, researchers have begun to apply Deep Neural Networks (DNN), Convolutional Neural Networks (CNN), and Recurrent Neural Networks (RNN) to SER tasks. In particular, Long Short-Term Memory (LSTM) networks, with their unique gating mechanisms that effectively capture long-term dependencies in time-series data, have become the mainstream approach in emotion recognition.

However, single-layer LSTM structures still have limitations in extracting emotional features, especially when dealing with audio signals with mixed or complex emotional shifts. To address this, we introduced an additional LSTM layer to form a dual-layer LSTM model that can better capture and interpret emotional information, thereby improving accuracy and efficiency in emotion recognition tasks\cite{b1}.

\section{Related Work}
Emotion recognition technology initially relied heavily on handcrafted features. These features typically contain noisy and irrelevant information, which will impede the model's performance. Fuzzy rough sets, as a powerful mathematical framework for measuring fuzzy relation and uncertainty, has been widely used in feature selection, soft computing, and medical diagnosis. Gao et al. \cite{gao2022parameterized} proposed a fuzzy rough sets-based attribute reduction method. In \cite{xing2022weighted}, a weighted fuzzy rough sets classifier was proposed for semi-supervised learning and has been applied in medical diagnosis, achieving desirable results. Despite the improvements they achieved, one key limitation is overlooked however, that is, these handcrafted features often struggle to fully express the complexity of emotions and become inefficient when processing large datasets.

With the rise of deep learning, automatic feature extraction has gradually replaced traditional feature engineering. Models such as Convolutional Neural Networks (CNN) and Recurrent Neural Networks (RNN) are widely used in SER tasks. CNN’s convolutional structure can capture local features in audio data, making it suitable for extracting short-term emotional information, but it has limitations in capturing long-term dependencies. In contrast, LSTM networks, due to their unique gating structure, effectively handle long-term dependencies in time series and perform well in speech recognition and emotion analysis tasks.

Studies have shown that multi-layer LSTM structures exhibit greater feature extraction capabilities for processing complex emotional information. By stacking LSTM layers, models can extract and process emotional features layer by layer, thus enhancing emotion recognition capabilities. However, excessive LSTM layers can increase computational overhead and lead to gradient vanishing or exploding problems, affecting model training. Therefore, selecting the appropriate number of layers is crucial for improving model performance. This study adds an additional layer to the single-layer LSTM to form a dual-layer structure, verifying its effectiveness in emotion recognition tasks.

\section{Methodology}
\subsection{Data Preprocessing}
This study uses the RAVDESS dataset as the primary data source for training and testing. The Ryerson Audio-Visual Database of Emotional Speech and Song (RAVDESS) contains audio files with various emotional categories (such as anger, happiness, sadness), providing rich emotional features. To ensure data consistency and model effectiveness, several preprocessing steps were performed:
\begin{itemize}
    \item Mono Conversion: All audio samples were converted to mono to reduce processing complexity and ensure uniformity of input format.
    \item Resampling: The sampling rate of all audio data was standardized to 16 kHz, which better suits the model’s input requirements and improves processing efficiency.
    \item MFCC Feature Extraction: Mel-Frequency Cepstral Coefficients (MFCC) were extracted as input features, which effectively capture frequency distribution and emotional information in speech signals. Each audio sample was converted into 20 time frames, each with 40-dimensional MFCC features as model inputs.
\end{itemize}

\subsection{Model Architecture}
Based on the existing single-layer LSTM model, this study proposes a dual-layer LSTM structure to further enhance the capture of emotional features in audio signals. The architecture of the dual-layer LSTM model is designed as follows:
\begin{itemize}
    \item Input Layer: Preprocessed audio data is converted into a sequence of feature vectors, with a sequence length of 20 and a feature dimension of 40 for each sample.
    \item Dual-layer LSTM: The first LSTM layer initially extracts emotional features from the audio signal, while the second LSTM layer further processes and enriches the emotional information, enhancing the understanding of emotional transitions and long-term dependencies. Each LSTM layer has 128 hidden units to ensure expressive power and computational efficiency.
    \item Fully Connected Layer: The output of the LSTM layers is flattened and passed to the fully connected layer, which maps emotional features to target emotion categories.
    \item Softmax Output Layer: The model outputs a probability distribution of each emotion category through the Softmax activation function, completing the emotion classification task.
\end{itemize}

\subsection{LSTM Network Architecture and Mathematical Derivation}

Long Short-Term Memory (LSTM) networks are a special type of Recurrent Neural Network (RNN) introduced by Hochreiter and Schmidhuber to address the vanishing and exploding gradient problems in traditional RNNs \cite{hochreiter1997long}. LSTM networks are widely used in emotion recognition from speech data because of their ability to model long-term dependencies, which are crucial for capturing temporal emotional patterns. The core of an LSTM lies in its gating mechanisms, which consist of the input gate, forget gate, and output gate. These gates dynamically control the flow of information and efficiently handle long-term dependencies.

\paragraph{Components of an LSTM Cell}

An LSTM cell consists of four main components: the cell state, input gate, forget gate, and output gate. At each time step, the LSTM updates the cell state based on the current input and the previous hidden state, enabling the network to selectively retain or forget information from past time steps.

1. \textbf{Forget Gate}: The forget gate decides what portion of the previous cell state, \( C_{t-1} \), should be retained. It utilizes a sigmoid activation function, outputting a value between [0, 1] to control the retention of previous information.
   \begin{equation}
       f_t = \sigma(W_f \cdot [h_{t-1}, x_t] + b_f)
   \end{equation}
   where \( W_f \) and \( b_f \) are the weight matrix and bias for the forget gate, \( h_{t-1} \) is the previous hidden state, and \( x_t \) is the current input.

2. \textbf{Input Gate}: The input gate decides what portion of the new information should be stored in the cell state. It consists of a sigmoid function controlling the amount of new information and a tanh activation function generating the candidate cell state \( \tilde{C}_t \).
   \begin{equation}
       i_t = \sigma(W_i \cdot [h_{t-1}, x_t] + b_i)
   \end{equation}
   \begin{equation}
       \tilde{C}_t = \tanh(W_c \cdot [h_{t-1}, x_t] + b_c)
   \end{equation}

3. \textbf{Cell State Update}: The cell state is updated by combining the previous cell state and the new candidate cell state. The forget gate controls how much of the previous cell state is retained, while the input gate determines how much of the new information is added.
   \begin{equation}
       C_t = f_t \ast C_{t-1} + i_t \ast \tilde{C}_t
   \end{equation}
   Here, \( \ast \) represents the element-wise multiplication. The updated cell state \( C_t \) thus retains relevant historical information while integrating the current input.

4. \textbf{Output Gate}: The output gate generates the current hidden state \( h_t \), which is passed to the next time step and serves as the final output for classification tasks. The output gate uses a sigmoid function to control the flow of the cell state \( C_t \), activated by a tanh function.
   \begin{equation}
       o_t = \sigma(W_o \cdot [h_{t-1}, x_t] + b_o)
   \end{equation}
   \begin{equation}
       h_t = o_t \ast \tanh(C_t)
   \end{equation}
   where \( h_t \) is the hidden state used for subsequent time steps as well as the final output for emotion classification.

\paragraph{Flow of Information in LSTM Networks}

LSTM networks operate with two primary flows of information: the cell state \( C_t \) and the hidden state \( h_t \). The cell state is propagated across time steps, preserving essential information and selectively forgetting irrelevant information through the forget gate. The hidden state, on the other hand, is used to generate output at each time step, making LSTM highly effective in sequence-based tasks such as speech emotion recognition.

The LSTM architecture introduced by Hochreiter and Schmidhuber \cite{hochreiter1997long} has become foundational for sequential modeling tasks. Gers et al. \cite{gers2000learning} proposed the Peephole LSTM variant to further refine cell state control, enhancing the model's adaptability to complex time-series data. Recent studies, such as Heigold et al. \cite{heigold2016end}, have validated the effectiveness of LSTM for speech emotion recognition tasks, demonstrating its ability to capture hierarchical emotional features. Multi-layer LSTM structures, as investigated in our study, allow for stepwise extraction of high-level emotional features, leading to improved recognition accuracy in complex emotional patterns.

\subsection{RAVDESS Dataset Overview}

The Ryerson Audio-Visual Database of Emotional Speech and Song (RAVDESS) is a standardized multimodal dataset specifically designed for research in affective computing, emotion recognition, and multimodal human-computer interaction. Developed by Ryerson University, the dataset includes high-quality audio and video recordings produced by professional actors under controlled laboratory conditions. Due to its diverse set of emotional categories and intensity levels, the RAVDESS dataset is widely used for evaluating the performance of speech and video-based emotion recognition systems.

\subsection{Dataset Structure}

The RAVDESS dataset is organized into two main parts: \textit{Speech Emotions} and \textit{Song Emotions}, each containing multiple emotional categories and corresponding intensity levels:

\begin{itemize}
    \item \textbf{Speech Emotions}: 
    \begin{itemize}
        \item Comprises 1,440 audio files, recorded by 24 actors (12 male and 12 female) expressing 8 distinct emotions: neutral, anger, fear, sadness, happiness, disgust, surprise, and calm.
        \item Each emotion, except for neutral, is presented in two intensity levels: normal and strong, adding nuance to the emotional portrayal.
        \item Each audio file is in WAV format with a 48kHz sampling rate, and the average duration is approximately 3 seconds, ensuring clarity and consistency of emotional expression.
    \end{itemize}
    \item \textbf{Song Emotions}:
    \begin{itemize}
        \item Contains 1,012 singing files where the same 24 actors sing short melodies with different emotional tones.
        \item Song emotions follow similar categories and intensities as the speech emotions, suitable for examining emotional expression in singing versus spoken speech.
        \item The singing files are also in WAV format, providing high-quality audio suitable for detailed analysis.
    \end{itemize}
\end{itemize}

\subsection{Data Instances}

Each audio file in the dataset is labeled with a structured filename encoding essential information about the emotion, gender, and intensity. For example, the filename \texttt{03-01-05-01-02-01-12.wav} represents the following details:
\begin{itemize}
    \item \texttt{03}: Emotion category (e.g., happiness).
    \item \texttt{01}: Emotion intensity (normal intensity).
    \item \texttt{02}: Gender of the actor (male).
    \item \texttt{12}: Actor ID.
\end{itemize}
This naming convention allows researchers to quickly identify the emotion category and other attributes of each audio file, facilitating efficient data loading and labeling.

\subsection{Dataset Usage}

The RAVDESS dataset is widely utilized in emotion recognition tasks, often serving as the primary source of training and testing data. Due to its richness in emotional diversity and audio quality, this dataset supports the development and evaluation of various emotion recognition models. The standard usage steps for the RAVDESS dataset include:

\begin{enumerate}
    \item \textbf{Data Loading and Preprocessing}: Read audio files and convert them to suitable audio formats and sampling rates. Common preprocessing steps include converting to mono, resampling, and normalization.
    \item \textbf{Feature Extraction}: Extract features from audio files, such as Mel Frequency Cepstral Coefficients (MFCC), chroma features, or pitch features, which are then used for emotion modeling.
    \item \textbf{Label Generation}: Utilize the encoded information in the filenames to extract emotion labels and intensity, creating labels for classification tasks.
    \item \textbf{Model Training and Testing}: Input processed audio features into the model and evaluate emotion classification based on the labels.
\end{enumerate}

The RAVDESS dataset provides high-quality multimodal emotional samples, and it can be downloaded from the \textit{official website} \cite{ravdess}. In practice, the RAVDESS dataset is often used in combination with other emotional datasets, such as Emo-DB, to test model generalizability across different data sources.

\subsection{Training Process}
The cross-entropy loss function was used in this experiment, and the Adam optimizer was used for training. The initial learning rate was set to 0.001 and gradually decayed during training to ensure model convergence. To prevent overfitting during training, learning rate was applied to the LSTM layers, with a dropout rate of 0.001. The model was trained for 60 epochs on a GPU to accelerate computation.

\section{Experiments}
\subsection{Experimental Setup}
The experiments were conducted on the RAVDESS dataset, with an 80:20 random split for the training and test sets. The main parameters of the model training are set as follows:
\begin{itemize}
    \item Learning rate: 0.001
    \item Batch size: 64
    \item Training epochs:60
\end{itemize}
By maintaining consistent experimental parameters, we ensured the reliability of the training and test results.

\subsection{Performance Evaluation Metrics}
To comprehensively evaluate model performance, this experiment uses accuracy as evaluation metrics. Accuracy measures the overall classification performance, while recall and F1-score assess the model’s ability to recognize each emotional category. The average computation time reflects the model’s real-time processing capability.

\subsection{Experimental Results}
TThe confusion matrix illustrating the experimental results is shown as\ref{fig-1} below. Our system performed exceptionally well on the small RAVDESS dataset. It achieved a relatively high recognition rate across all eight emotion labels in this dataset, with an overall average recognition rate of 87.33\%.
\begin{figure}[htbp]
\centerline{\includegraphics[width=1.2\linewidth]{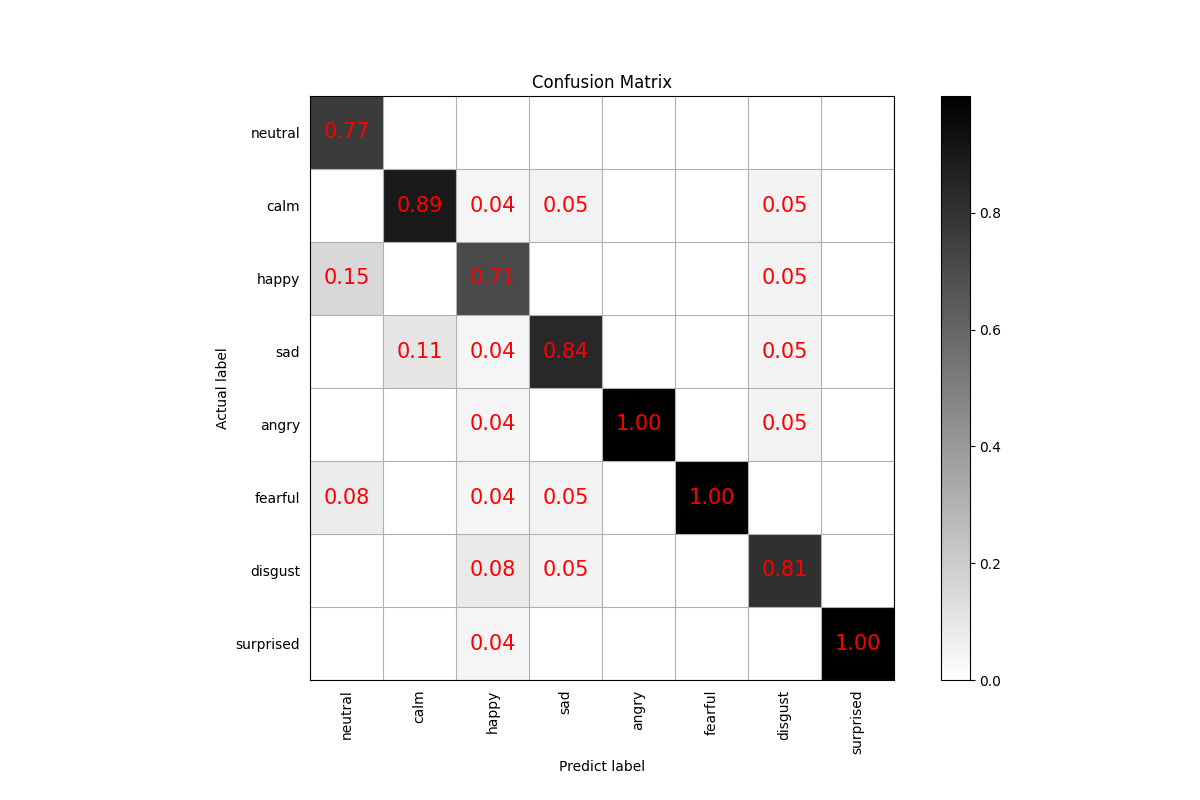}}
\caption{Confusion matrix of the experimental results}
\label{fig-1}
\end{figure}

\section{Results and Discussion}
The experimental results indicate that adding an additional LSTM layer significantly enhances the model’s emotion recognition ability. Compared to the single-layer structure, the dual-layer LSTM improved classification accuracy by 2\% and reduced average processing time. This architecture strengthened the model's ability to capture complex emotional patterns, enabling it to better understand emotional features in audio data. Particularly, the dual-layer structure excels over the single-layer structure in extracting long-term emotional dependencies.

While the dual-layer LSTM structure achieved notable results, the added layer also increased computational cost and memory consumption. Future research may explore techniques such as model pruning and quantization to reduce resource consumption and improve model efficiency in real-time applications. Additionally, integrating other sequence modeling methods, such as Transformer or hybrid model architectures, may further improve emotion recognition performance.

\section{Conclusion}
Based on an existing SER model, this study introduces a dual-layer LSTM structure, significantly enhancing the model’s emotion recognition accuracy and processing efficiency. The dual-layer structure improves the model’s ability to capture complex emotional features, making it more adaptable for diversified emotion recognition tasks. The experimental results demonstrate the effectiveness of dual-layer LSTM in handling long-term dependencies, effectively improving classification performance and real-time capabilities. Future research can further optimize computational resource usage while maintaining performance and explore the integration of multiple sequential modeling techniques to enable wide deployment of emotion recognition systems across various applications.
\section*{Acknowledgment}

This study extends special thanks to our lab colleagues for their support in data processing and experimental design. Additionally, we appreciate the contributors in the open-source community for developing and sharing foundational models and tools, which provided essential references for our experiments. Finally, we are grateful to the developers of the RAVDESS dataset for providing high-quality emotional data, enabling emotion recognition research to be validated on real-world data.

\vspace{12pt}

\end{document}